\documentclass[conference]{IEEEtran}
\IEEEoverridecommandlockouts

\usepackage{cite}
\usepackage{multirow}
\usepackage{amsmath,amssymb,amsfonts}
\usepackage{algorithmic}
\usepackage{graphicx}
\usepackage{textcomp}
\usepackage{xcolor}
\usepackage{booktabs}
\usepackage{algorithm}
\usepackage{subcaption}
\usepackage{makecell}    
\usepackage{cleveref}

\def\BibTeX{{\rm B\kern-.05em{\sc i\kern-.025em b}\kern-.08em
    T\kern-.1667em\lower.7ex\hbox{E}\kern-.125emX}}

\begin{document}

\title{Are All Data Necessary? Efficient Data Pruning for Large-scale Autonomous Driving Dataset via Trajectory Entropy Maximization\\


\author{
    Zhaoyang Liu, Weitao Zhou, Junze~Wen, Cheng Jing, Qian Cheng, Kun Jiang and Diange Yang
    \thanks{Zhaoyang Liu, Weitao Zhou, Junze~Wen, Cheng Jing, Qian Cheng, Kun Jiang and Diange Yang are with the School of Vehicle and Mobility, Tsinghua University, Beijing~100084,~China 
}
    \thanks{Corresponding authors: Weitao Zhou and Diange Yang 
    ({\tt\small zhouwt@tsinghua.edu.cn}, {\tt\small ydg@tsinghua.edu.cn}).}
}

}


\maketitle

\begin{abstract}
Collecting large-scale naturalistic driving data is essential for training robust autonomous driving planners. However, real-world datasets often contain a substantial amount of repetitive and low-value samples, which lead to excessive storage costs and bring limited benefits to policy learning.
 To address this issue, we propose an information-theoretic data pruning method that effectively reduces the training data volume without compromising model performance. Our approach evaluates the trajectory distribution information entropy of driving data and iteratively selects high-value samples that preserve the statistical characteristics of the original dataset in a model-agnostic manner.
 From a theoretical perspective, we show that maximizing trajectory entropy effectively constrains the Kullback-Leibler divergence between the pruned subset and the original data distribution, thereby maintaining generalization ability.
 Comprehensive experiments on the NuPlan benchmark with a large-scale imitation learning framework demonstrate that the proposed method can reduce the dataset size by up to 40\% while maintaining closed-loop performance.
 This work provides a lightweight and theoretically grounded approach for scalable data management and efficient policy learning in autonomous driving systems. 
\end{abstract}


\section{Introduction}
\label{section:introduction}

Autonomous driving has achieved remarkable progress in recent years, powered by large-scale naturalistic driving data and deep learning. These data-centric systems continuously learn from fleet-level experiences to improve perception, planning, and control \cite{li2023survey, wang2024survey, zhou2022dynamically}. However, as autonomous fleets expand, the challenge has shifted from data scarcity to data efficiency—the ability to identify and retain only those samples that meaningfully enhance model performance \cite{cao2023continuous }.

 Leading companies such as Tesla and Waymo collect vast amounts of real-world data daily, yet only a small portion corresponds to rare, safety-critical, or long-tail scenarios\cite{zhou2022long}. Most trajectories are repetitive and low-risk, leading to redundant data accumulation. This imbalance inflates storage, annotation, and retraining costs while providing diminishing further performance gains\cite{kaplan2020scaling, sorscher2022beyond}. As a result, determining which data are truly valuable before model training has become a critical bottleneck in achieving scalable and sustainable autonomous driving.

Previous efforts to identify high-value data have followed several directions. Heuristic or scenario-based approaches rely on disengagement logs, collision triggers, or anomaly events to tag important samples, but their effectiveness is limited by manually chosen thresholds and scenario definitions. Model-based valuation methods—such as Shapley value estimation or gradient-based importance scoring—evaluate each sample’s contribution to model performance but demand costly retraining or gradient computations. Although these techniques have deepened the understanding of data quality, they remain difficult to scale due to substantial computational cost and dependence on specific model architectures.

\begin{figure}[t]
  \centering
  \includegraphics[width=\linewidth]{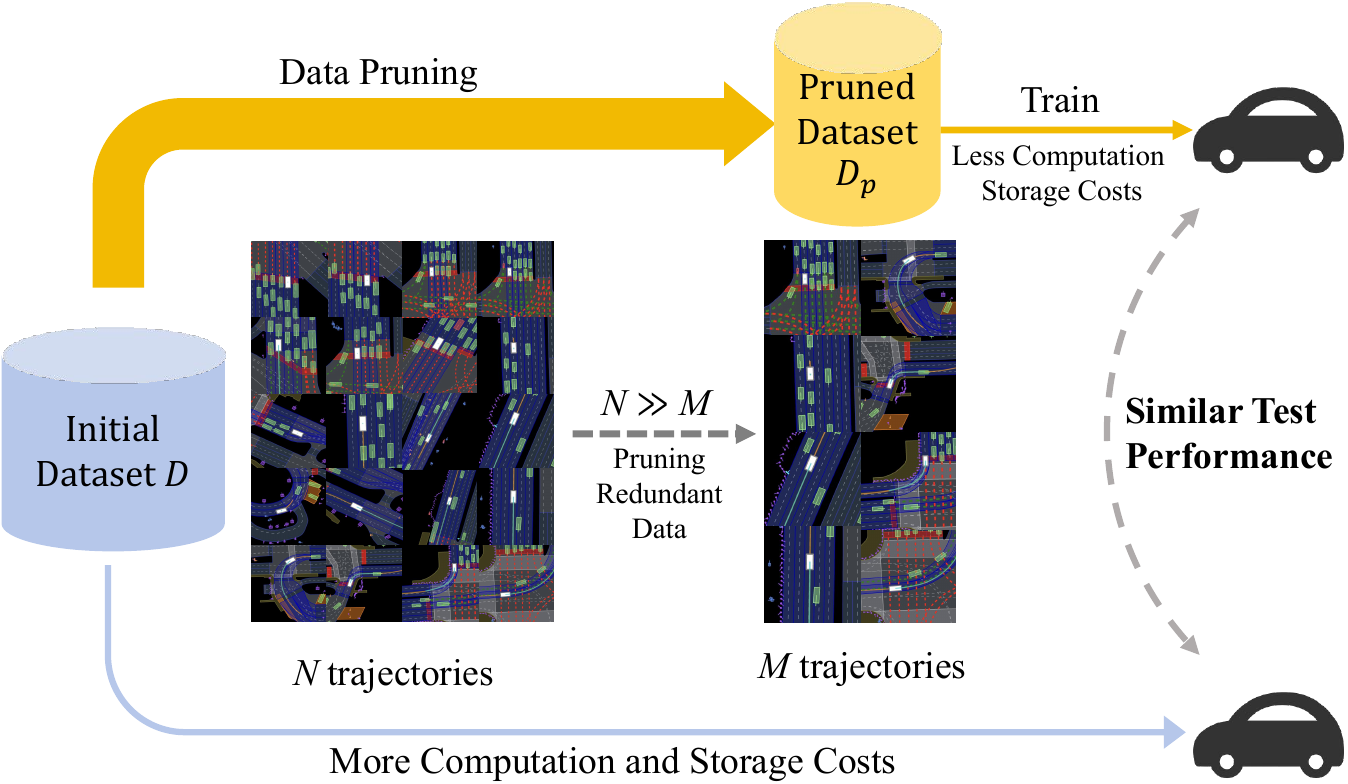} 
  \label{fig:problem}
  \caption{In autonomous driving, vast amounts of naturalistic driving data are collected from real-world operations. However, only a small subset contributes meaningfully to model improvement, while most samples are redundant. Our goal is to automatically identify and retain high-value data in an online and scalable manner, reducing storage and training costs without compromising performance} 
\end{figure}

To address these challenges, we propose a lightweight, model-agnostic framework for autonomous driving data pruning that directly evaluates data value from intrinsic trajectory statistics. We introduce the concept of trajectory-distribution entropy, providing a principled measure for assessing scene diversity and representativeness without relying on predefined rules or model retraining. By maximizing entropy coverage, our approach removes redundant trajectories while preserving the dataset’s overall distribution. This enables scalable and potentially online deployment on vehicle fleets to filter uninformative data prior to centralized training. Experiments on the NuPlan benchmark show that the proposed method can reduce dataset size by about 40\% with minimal performance degradation, demonstrating its practicality for scalable, efficient, and distribution-preserving data pruning in autonomous driving.

 The main contributions are summarized as follows:
The main contributions of this paper are summarized as follows: 

\begin{itemize} 
\item We formalize trajectory-distribution entropy as a measure of information diversity within driving trajectories, enabling data valuation based on intrinsic scenario variety as a practical criterion for assessing data value in autonomous driving; 
\item We design a low-complexity, model-agnostic pruning algorithm that efficiently identifies redundant data while preserving scene and trajectory diversity, enabling scalable or online deployment; 
\item We validate our approach on NuPlan, showing that it achieves up to 40\% data reduction without degrading driving policy performance, greatly improving scalability and efficiency. 
\end{itemize} 

\section{Related Works}
\label{section:related_work}
Research on efficient data utilization in autonomous driving can be broadly divided into three lines: heuristic and feature-based methods, model-based methods, and dataset distillation methods.

\subsection{Heuristic and feature-based Methods}

Current industrial practice approached data selection as a rule-design problem—identifying valuable or safety-critical moments through heuristic filters such as disengagement events, collisions, or high-risk maneuvers\cite{zhou2025drarl}.  Other projects construct curated long-tail subsets by identifying corner-case driving situations, either through sensor triggers or post-hoc clustering of trajectory statistics \cite{hallgarten2024can, hao2025driveaction}. 

For example, Li Auto's SUP-AD\cite{tian2024drivevlm} mines long tail or challenging scenarios from large autonomous driving databases, and then manually filters them to obtain multiple datasets with challenging long tail scenarios. Waymo's WOD-E2E\cite{xu2025wode2ewaymoopendataset} utilizes rule-based heuristics and multimodal large language model to mine data from massive raw datasets, constructing a highly rare long tail scene dataset. Tesla's "shadow mode" identifies scenarios with a divergence between the model's planned trajectory and the driver's ground-truth actions, marking them as key data for storage. \cite{mahmud2017application} rely on purely rule-based methods, such as Time-to-Collision (TTC) and Crash Index (CI), while \cite{zhang2023perception} categorize scenario datasets based on weather conditions.

While these methods highlight the practical relevance of heuristic screening, they remain constrained by manual criteria and predefined thresholds. Consequently, their generalization to unanticipated edge cases or evolving traffic patterns is limited, and maintaining consistent standards across fleets demands significant human effort.



\subsection{Model-Based Data Valuation}

Model-based approaches formulate data selection as a feedback problem in which the contribution of each sample is estimated from its effect on model learning. Typical criteria include gradient- and error-based sensitivities, training-trajectory tracing, and Shapley-style or influence-function attributions. 

For instance, inspired by the phenomenon of catastrophic forgetting, \cite{toneva2018empirical} introduce the concept of "forgetting events" to distinguish datasets by examining how easily different examples are forgotten. \cite{pmlr-v97-ghorbani19c} propose a data pruning method based on Shapley values, a concept from game theory, which quantifies the contribution of each data point. This is approximated by analyzing the gradients corresponding to individual data points during training. Similarly, \cite{paul2021deep} estimate data value using Gradient Norm (GraNd) and Error L2-Norm (EL2N) scores for pruning, while \cite{yang2022dataset} assess value by estimating the influence of data points on model parameters. In a different approach, \cite{evans2024bad} frame the problem within an active learning paradigm; by training teacher and student models, they quantify data value as "learnability", defined by the performance gap between the two models on a given data point.

These techniques strengthen the linkage between data and downstream metrics, but they assume access to the model (and often its gradients) and frequently require repeated optimization passes or ensembles. Under fleet-scale, continuously collected streams, such requirements impose prohibitive latency and compute, rendering model-dependent valuation unsuitable for real-time or pre-training filtering, particularly on-vehicle.



\subsection{Dataset Distill Methods}
Dataset distillation methods aim to construct compact, information-preserving datasets that approximate the learning behavior of large-scale corpora. These approaches optimize a small set of synthetic or representative samples to reproduce the gradients, feature distributions, or optimization trajectories of models trained on full datasets. 

For example, \cite{wang2018dataset} synthesize a small dataset through optimized algorithms for simple image classification. After training on this dataset, the model can quickly approach the accuracy of the model trained on the complete dataset. The problem lies in its strong dependence on model initialization weights, poor robustness and lack of interpretability in synthesizing data, and the huge computational cost of synthesizing small datasets, which is completely unsuitable for us to handle large-scale autonomous driving datasets. In contrast, \cite{cazenavette2022dataset} align the gradient updates of a student model with those of a teacher model on real data, enabling the synthesis of a small dataset that yields comparable performance. However, unlike data pruning, the shape of the synthesized data is completely different from the original data. \cite{yin2023squeeze} use graph neural networks to capture semantic information between samples and enhances diversity through adversarial perturbations, but the resulting compressed dataset is actually an equivalent dataset with some information loss. The key information of the original data is actually transplanted into the trained model.

Nevertheless, transferring dataset distillation techniques to autonomous driving remains non-trivial. The primary objective of dataset distillation is to compress a dataset by extracting its core information into a smaller, often synthetic, set. Consequently, the original data format is altered, inevitably leading to a loss of human interpretability and raw information fidelity, and this method is not suitable for autonomous driving datasets. Furthermore, most distillation methods are designed for image classification, relying on the assumption that data samples are independent and identically distributed (i.i.d.) while driving data are sequential, interactive, and safety-critical, fundamentally violating this core assumption.



\begin{figure}[t]
  \centering
  \includegraphics[width=\linewidth]{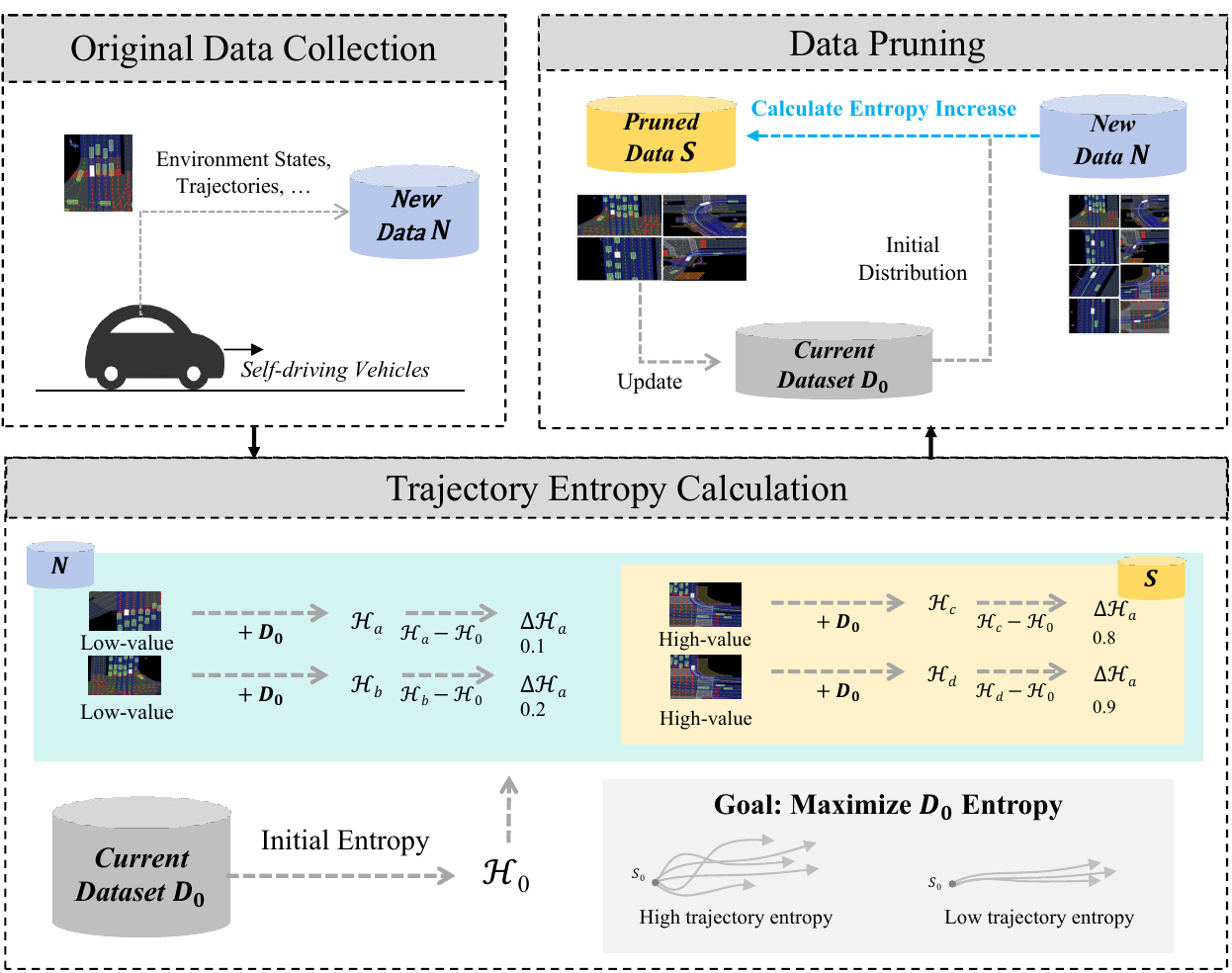} 
  \label{fig:prun}
  \caption{The framework of our method: by calculating the trajectory distribution information entropy, we determine whether the new data on the vehicle side can increase the information entropy value relative to the existing data, thereby achieving fast and efficient data pruning.} 
\end{figure}

\section{Method}
\label{section:methodology}

This paper presents the proposed methodology through four core components: problem formulation, the idea of trajectory-distribution information entropy, mathematical discourse of entropy maximization, and the design of the data pruning algorithm.  

\subsection{Problem Formulation}

In the context of the closed-loop data lifecycle of autonomous driving~\cite{li2023survey, wang2024survey}, we construct the data pruning problem within a unified framework that reflects real-world development and deployment processes. The closed-loop process typically consists of three stages: data collection, model training, and model deployment with testing and refinement.

\paragraph{Data collection.} In this stage, the autonomous driving system continuously acquires large-scale perception and decision data from vehicle fleets. Each dataset $D$ can be expressed as $D = {S_i, T_i, i = 1, \ldots, n}$, where $S_i$ denotes perception features and $T_i$ represents the corresponding driving trajectory. The perception features include ego-vehicle states, HD map $M$, static obstacles $N_S$, dynamic agents $N_A$, and traffic context $C$ such as lights and signals. The decision trajectory $T_i$ is a sequence of future waypoints ${x_i, y_i, \alpha_i, i = 0, \ldots, t}$, where $(x_i, y_i)$ denote the spatial position and $\alpha_i$ the heading angle in the ego-centric coordinate frame. For simplicity, the current pose $(x_0, y_0, \alpha_0)$ is set to $(0, 0, 0)$.

\paragraph{Model training.} Given dataset $D$, the $\mathrm{Model}$ parameterized by $\theta$ is trained to predict the future trajectory $\overline{T}i = \mathrm{Model}(S_i; \theta)$. $\theta = \mathrm{Train}(D)$ denotes the model parameters obtained from training on dataset D. The performance of the model is evaluated through a testing process $\mathrm{Test}(\cdot)$ over unseen scenarios $S_{\mathrm{test}}$, producing a score:
\begin{equation}
R_{\mathrm{test}} = \mathrm{Test}(S_{\mathrm{test}}, \mathrm{Model}(S_{\mathrm{test}}; \mathrm{Train}(D))).
\end{equation}

\paragraph{Data pruning objective.} The aim of pruning is to derive a reduced dataset $D_p = \mathrm{Pruning}(D)$, where $D_p = {S_i, T_i, i = 1, \ldots, n_p}$ and $n_p < n$. The pruning ratio is defined as $(1 - n_p/n) \times 100\%$. The optimization objective can be expressed as maximizing this ratio while maintaining comparable performance to the original model:
\begin{equation}
R_p = \mathrm{Test}(S_{\mathrm{test}}, \mathrm{Model}(S_{\mathrm{test}}; \mathrm{Train}(D_p))) \approx R_{\mathrm{test}}.
\end{equation}

\subsection{Trajectory Distribution Information Entropy}

A major challenge in autonomous driving is the mismatch between the training and deployment distributions. During real-world testing, the vehicle frequently encounters scenarios not covered by the training data, resulting in trajectories that fall outside the learned manifold and degrade performance. To achieve stable generalization, it is therefore crucial to maintain the statistical diversity of the training data after data pruning and preserve the essential characteristics of the original distribution.

While the overall distribution of an autonomous driving dataset is complex, the trajectory data (of ego vehicle) within it is relatively structured and compact. Driving can be viewed as a transformation from high-dimensional perception inputs to low-dimensional trajectory outputs, where the environmental diversity is directly reflected in the trajectory space. In this work, we take the distribution of trajectories as a representation for scenario diversity in the data.

This work defines the concept of trajectory distribution information entropy, inspired by the entropy formulations in statistical mechanics and maximum entropy reinforcement learning\cite{berrueta2024maximum}. Entropy, as a measure of uncertainty and diversity, captures how evenly data are distributed across possible outcomes. In the context of autonomous driving, we extend this notion to quantify the diversity and representativeness of trajectory distributions in large-scale driving datasets, enabling a principled evaluation of data coverage and balance. 

Given a trajectory dataset $\mathcal{T} = \{T_i\}_{i=1}^n$, where each trajectory $T_i = \{x_i, y_i, \alpha_i\}_{t=0}^T$ represents ego-vehicle motion in a local coordinate frame, we discretize the spatial domain into a 3D grid $(X, Y, \alpha)$ with minimum cell resolution $\delta$. Denote $N_i$ as the number of trajectory samples in grid cell $i$ (considering only $N_i > 0$) and $N$ as the total number of trajectory points. The trajectory distribution information entropy is defined as:
\begin{equation}
H = - \sum_i \frac{N_i}{N} \log \frac{N_i}{N}.
\label{eq:entropy}
\end{equation}
Here, $H$ measures the overall statistical coverage of the dataset in trajectory space. In practice, we use two-dimensional projection $(X, Y)$ for computational efficiency, omitting $\alpha$ due to hardware and timing constraints.

\subsection{Data Pruning via Entropy Maximization}

Autonomous driving datasets are inherently imbalanced: rare long-tail scenarios appear infrequently but are crucial for model robustness and safety. Let the empirical trajectory distribution of the original dataset be $P = \{P_i\}_{i=1}^M$ and that of a pruned subset be $Q = \{Q_i\}_{i=1}^M$, where $i$ indexes discrete trajectory bins. Their Kullback–Leibler (KL) divergence is:
\begin{equation}
D_{\mathrm{KL}}(P \,\|\, Q) = \sum_{i=1}^M P_i \log \frac{P_i}{Q_i}.
\end{equation}
If any rare but important cell with $P_i > 0$ is omitted ($Q_i = 0$), $D_{\mathrm{KL}}$ diverges. To keep the divergence bounded, we require:
\begin{equation}
Q_i > 0, \quad \forall i \text{ s.t. } P_i > 0.
\end{equation}

Directly enforcing this constraint is impractical when $P$ is unknown or continuously evolving. Instead, we adopt a surrogate objective by maximizing the entropy of $Q$:
\begin{equation}
H(Q) = -\sum_{i=1}^M Q_i \log Q_i.
\end{equation}
Maximizing $H(Q)$ implicitly encourages uniform coverage of the trajectory space. Since rare trajectories contribute disproportionately to entropy increase, this objective naturally prioritizes their retention. Denoting $\varepsilon > 0$ as the minimum probability assigned to any retained cell, we have:
\begin{equation}
D_{\mathrm{KL}}(P\|Q) = -H(P) - \sum_i P_i \log Q_i 
\le -H(P) - \sum_i P_i \log \varepsilon,
\end{equation}
ensuring $D_{\mathrm{KL}}$ remains finite. Therefore, maximizing the trajectory distribution entropy provides a principled way to bound distributional deviation while reducing data volume.

\subsection{Entropy-Guided Pruning Algorithm}

We aim to obtain a subset $\overline{T} \subset T$ such that the entropy of its trajectory distribution is maximized:
\begin{equation}
\overline{T} = \arg\max_{\overline{T} \subset T} H(\overline{T}).
\end{equation}
However, directly solving this optimization is computationally prohibitive and may yield excessively small subsets that degrade learning performance. To address this, we propose a lightweight incremental selection algorithm that approximates entropy maximization under a user-specified pruning ratio $\eta = (1 - n_p/n) \times 100\%$.

The procedure is as follows:  
(1) Randomly sample an initial subset $D_0$ and compute its entropy $H_0$;  
(2) Iteratively sample a batch of $N$ candidate trajectories and evaluate the entropy gain $\Delta H_i$ for each;  
(3) Select the top $(1-\eta)N$ samples with the highest $\Delta H_i$ and merge them into $D_0$;  
(4) Repeat until all data points have been considered.  
This yields a pruned dataset $D_p$ with approximately linear computational complexity $\mathcal{O}(n)$.

The proposed pruning method consists of the following steps in \cref{tab:algorithm}, where:
\begin{itemize}
    \item $D_0$: Initial subset (size $n_0$)
    \item $H_0$: Current subset entropy
    \item $N$: Batch size for candidate sampling
    \item $\eta$: Pruning ratio parameter ($0 < \eta < 1$)
    \item $\Delta H_i$: Entropy gain for sample $x_i$
\end{itemize}

\begin{algorithm}[!ht]
\caption{Entropy-Maximizing Data Pruning}
\begin{algorithmic}[1]
\STATE Initialize $D_0 \gets \text{RandomSample}(D, n_0)$
\STATE Compute $H_0 \gets \text{Entropy}(D_0)$
\WHILE{not all data sampled}
    \STATE $N \gets \text{RandomSample}(D \setminus D_0, N)$
    \FOR{each $x_i \in N$}
        \STATE $\Delta H_i \gets \text{Entropy}(D_0 \cup \{x_i\}) - H_0$
    \ENDFOR
    \STATE Sort $\{\Delta H_i\}$ in descending order
    \STATE Select top $(1-\eta) N$ samples: $S \gets \{x_i\}_{i=1}^{\eta N}$
    \STATE Update: $D_0 \gets D_0 \cup S$
    \STATE Update: $H_0 \gets \text{Entropy}(D_0)$
\ENDWHILE
\STATE \RETURN $D_0$
\end{algorithmic}
\label{tab:algorithm}
\end{algorithm}

The advantage of the above method is that its complexity is $\mathcal{O}(n)$ determined by the total amount of data $n$. The proposed method achieves a balance between statistical representativeness and computational efficiency. It ensures that the retained subset preserves the key statistical diversity of the original dataset, maintaining model performance while significantly reducing storage and training costs.

\section{Experiments}
\label{section:experiments}

\subsection{Experiment Setting}

We evaluate the proposed data pruning method on the NuPlan dataset~\cite{caesar2021nuplan}, a large-scale autonomous driving dataset from real-world environment. The Pluto model~\cite{cheng2024pluto} serves as our base policy, trained on 1M samples and evaluated on the Random14 benchmark. To emulate realistic data-rich conditions, the model size was reduced from 4.1M to 300k parameters, ensuring that performance variance originates from data selection rather than overcapacity. All experiments were conducted on a 4$\times$H20 workstation, with our pruning algorithm running on a single CPU.

Two modifications were applied to Pluto: (1) architecture compression to 300k parameters; and (2) removal of data augmentation to isolate the effect of data quality. We use the SPIDER tool for better develop efficiency \cite{qian2024spider} These adjustments ensure a clean comparison between unpruned, random-pruned, and entropy-pruned datasets.

\begin{figure}[t] 
  \centering
  \begin{subfigure}[t]{0.8\columnwidth}
    \includegraphics[width=\textwidth]{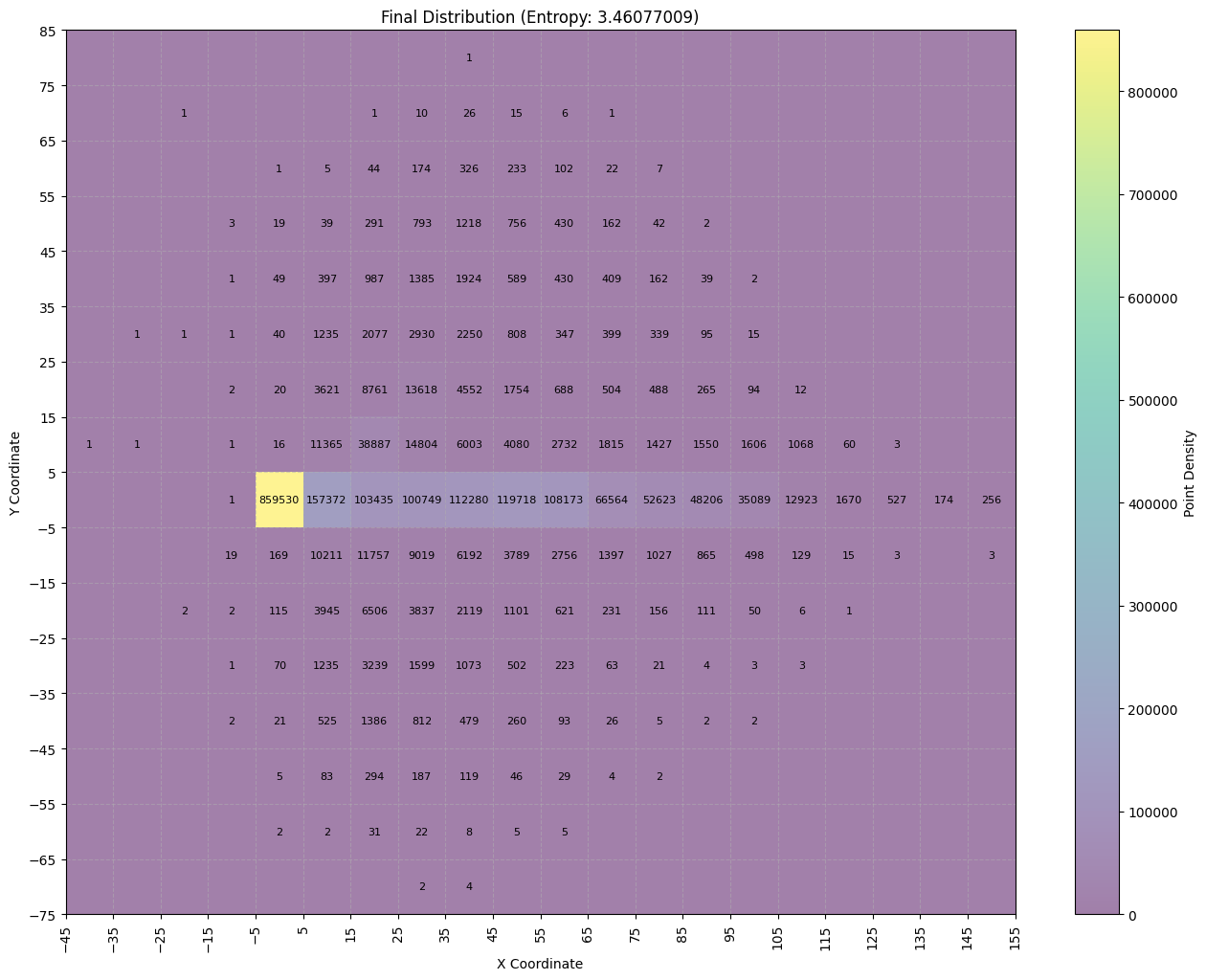} 
    \caption{The trajectory distribution information entropy of 1M NuPlan data in the experiments} 
    \label{fig:original}
    \vspace{0.5em} 
  \end{subfigure}
  \begin{subfigure}[t]{0.8\columnwidth}
    \includegraphics[width=\textwidth]{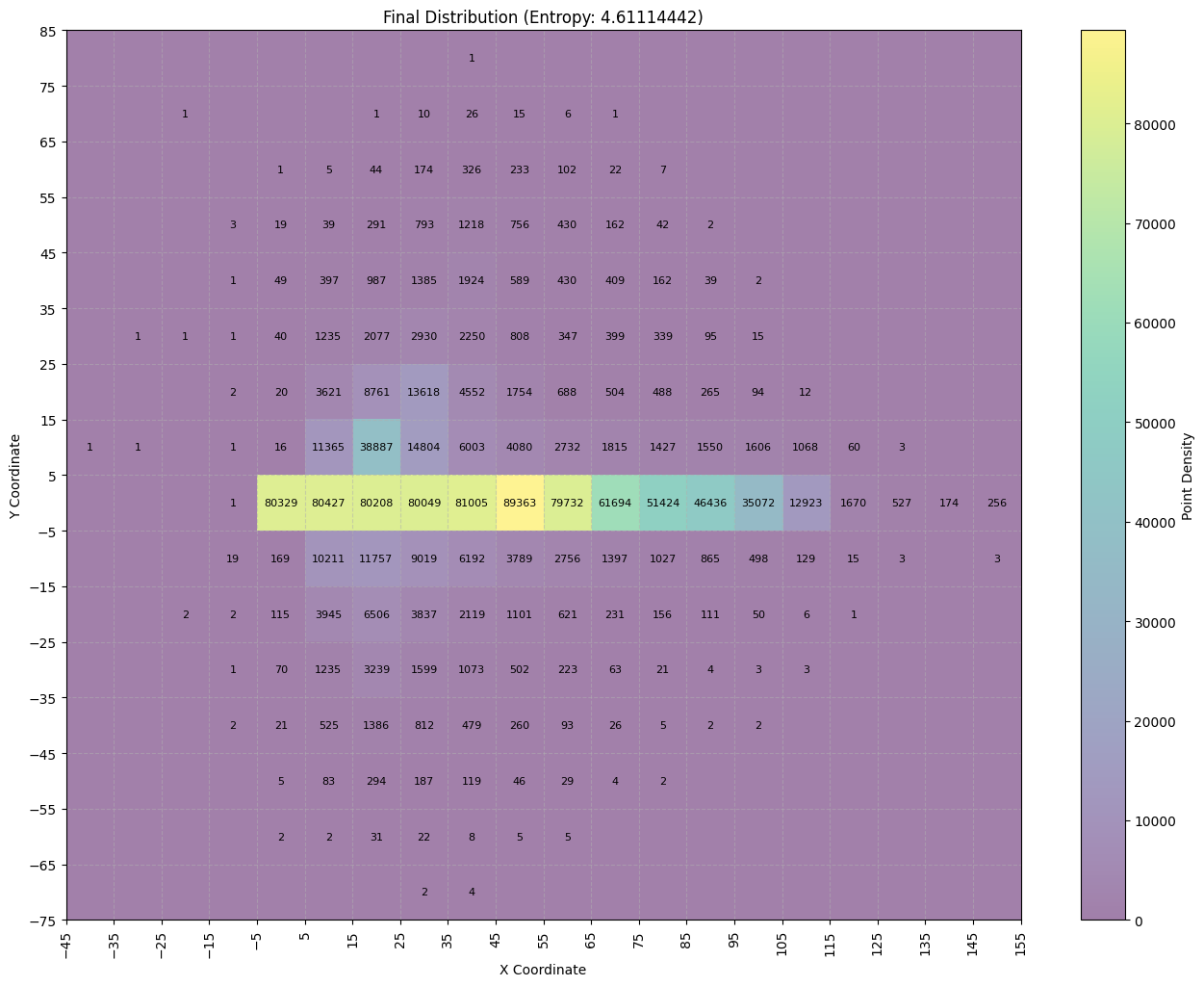} 
    \caption{The trajectory distribution information entropy of 50\% pruned NuPlan data with our pruning method in the experiments} 
    \label{fig:pruned}
    \vspace{0.5em}
  \end{subfigure}
  
  \caption{The visualization results of the trajectory distribution information entropy in the experiments show that the pruned entropy is higher. We can observe that our pruning method effectively removes a large number of repeated stationary and straight trajectories while preserving a smaller proportion of turning trajectories.}
  \label{fig:experimententropy}
\end{figure}

We conducted pruning experiments on the 1M dataset at varying pruning ratios (10\%, 20\%, 30\%, 40\%, 50\%) and compared model performance across multiple metrics. We show the results of trajectory distribution information entropy after pruning in \cref{fig:experimententropy}. 



\subsection{Baselines}

We compare our trajectory-entropy-based pruning with random pruning. Both methods share $\mathcal{O}(n)$ complexity and are model-agnostic, making random selection a fair reference. For each configuration, the model was trained to convergence and evaluated across multiple checkpoints to eliminate stochastic bias. 

The evaluation follows two expectations:  
(1) random pruning leads to monotonically degraded performance with increasing pruning ratio, and  
(2) entropy-based pruning maintains performance close to the unpruned baseline by selectively retaining informative trajectories.





\begin{table*}[t]
\centering
\caption{
Key metric comparison between entropy-based and random pruning on the NuPlan Random14 benchmark.
$\Delta$ denotes absolute improvement over random pruning.
Best values per pruning ratio are in \textbf{bold}.}
\label{tab:key_metrics_compact}
\setlength{\tabcolsep}{3pt}
\begin{tabular}{
c|
ccc|ccc|ccc|ccc|ccc|c
}
\toprule
\multirow{2}{*}{\textbf{Prune (\%)}} 
& \multicolumn{3}{c|}{\textbf{Score (↑)}} 
& \multicolumn{3}{c|}{\textbf{No Collisions (↑)}} 
& \multicolumn{3}{c|}{\textbf{Comfort (↑)}} 
& \multicolumn{3}{c|}{\textbf{Progress (↑)}} 
& \multicolumn{3}{c|}{\textbf{Drivable Comp. (↑)}} 
& \textbf{Ret. Gap (\%)} \\
\cmidrule(lr){2-4}\cmidrule(lr){5-7}\cmidrule(lr){8-10}\cmidrule(lr){11-13}\cmidrule(lr){14-16}
& Ours & Rand. & $\Delta$
& Ours & Rand. & $\Delta$
& Ours & Rand. & $\Delta$
& Ours & Rand. & $\Delta$
& Ours & Rand. & $\Delta$
& Ours–Rand. \\
\midrule
0 & 84.2 & 84.2 & 0.0  & 94.2 & 94.2 & 0.0  & 77.5 & 77.5 & 0.0  & 100.0 & 100.0 & 0.0  & 98.0 & 98.0 & 0.0  & 0.0 \\
10 & 82.6 & 82.6 & 0.0  & 92.4 & 94.2 & -1.8 & 82.0 & 79.1 & \textbf{+2.9} & 98.4 & 98.4 & 0.0  & 96.0 & 98.8 & -2.8 & 0.0 \\
20 & 82.6 & 82.0 & \textbf{+0.6} & 94.4 & 94.0 & \textbf{+0.4} & 77.1 & 77.9 & -0.8 & 98.8 & 99.6 & -0.8 & 97.6 & 98.0 & -0.4 & \textbf{+0.7} \\
30 & 83.5 & 74.7 & \textbf{+8.8} & 94.8 & 84.5 & \textbf{+10.3} & 84.3 & 75.9 & \textbf{+8.4} & 97.6 & 98.0 & -0.4 & 98.0 & 98.0 & 0.0  & \textbf{+10.5} \\
40 & 84.5 & 78.0 & \textbf{+6.5} & 94.5 & 89.0 & \textbf{+5.5}  & 84.7 & 77.9 & \textbf{+6.8} & 98.7 & 98.3 & \textbf{+0.4} & 99.2 & 97.2 & \textbf{+2.0} & \textbf{+7.8} \\
50 & 79.1 & 72.8 & \textbf{+6.3} & 89.1 & 84.5 & \textbf{+4.6}  & 81.9 & 78.3 & \textbf{+3.6} & 99.2 & 98.0 & \textbf{+1.2} & 97.6 & 95.2 & \textbf{+2.4} & \textbf{+7.5} \\
\bottomrule
\end{tabular}
\end{table*}

\subsection{Performance of Proposed Pruning Method}


As shown in \cref{tab:key_metrics_compact}, the proposed trajectory-entropy-based pruning method demonstrates strong robustness across all pruning ratios. Up to 40\% pruning, the overall score remains nearly identical to the full-dataset baseline (84.2~$\rightarrow$~84.5), confirming that entropy-guided selection preserves the statistical diversity and critical decision-making patterns required for generalization. Noticeable degradation appears only at 50\% pruning (79.1), where the data-to-parameter ratio approaches unity and the pruned set begins to under-sample the latent behavior space.

Across all metrics, our method consistently surpasses random pruning.
 In safety-related indicators such as No Collisions and Drivable Area Compliance, improvements of +5–10\% are observed at medium-to-high pruning levels (30–50\%), suggesting that entropy maximization preferentially retains rare yet safety-critical trajectories, effectively regularizing the data distribution.  Comfort and progress metrics also benefit, with smoother driving profiles (+8.4 at 30\%) reflecting reduced exposure to redundant straight-line scenes. Overall, the Retention Gap—defined as the relative performance retention with respect to the unpruned baseline—remains within +7–10\% at high pruning ratios, confirming that entropy-driven pruning achieves compression without compromising behavioral coverage. 

While maximizing entropy does not theoretically guarantee minimizing the KL divergence—nor does it enforce preservation of the original data distribution—the empirical evidence suggests that entropy-guided pruning remains highly effective in practice. By preferentially retaining diverse and low-frequency trajectory patterns, the method implicitly preserves the behavioral support needed for robust policy learning. This effect is particularly pronounced in safety-critical metrics such as \textit{No Collisions} and \textit{Drivable Area Compliance}, where entropy-based pruning consistently outperforms random sampling. Although the retained subset does not reproduce the original distribution, its broadened coverage in trajectory space appears sufficient to sustain model performance while eliminating redundant data. 


Random pruning causes progressive performance loss proportional to pruning severity, confirming that naive data reduction irreversibly discards critical information. The stable performance variance of our Pluto-mini baseline further rules out random fluctuations as a confounding factor.
The proposed method supports real-time, on-vehicle data filtering by computing trajectory distribution entropy with negligible computational overhead. This stands in contrast to active learning or model-driven pruning approaches, which require repeated model evaluations or full-dataset storage and are therefore impractical for fleet-scale continuous data collection.

Across a range of pruning ratios, the largest performance gains appear in safety-critical metrics such as \textit{No Collisions} and \textit{Drivable Area Compliance}. This aligns with the intended effect of entropy maximization: rare but safety-relevant trajectory patterns contribute disproportionately to entropy and are thus more likely to be retained. As a result, the pruned dataset preserves essential behavioral support even under substantial data reduction.

The model-agnostic nature of our approach illustrates that data-centric optimization can serve as an effective counterpart to conventional model-centric improvements. Rather than expanding data volume or model capacity, improving the statistical coverage of the training data offers a lightweight and scalable path toward performance retention.

Overall, the experiments demonstrate that entropy-based pruning is both an effective data curation mechanism and a practical system tool for autonomous driving pipelines. Its simplicity and computational efficiency make it particularly suitable for large-scale deployments, and future work will extend this framework to multi-modal sensor streams and adaptive, scenario-aware pruning schedules.

\section{Conclusions}
\label{section:conclusions}

In this paper, we tackle the challenge of dataset pruning in autonomous driving by introducing a trajectory distribution information-entropy-based method that selectively removes redundant data while preserving the overall distribution characteristics of the dataset. By maximizing the entropy of trajectory distributions, our approach retains the training efficacy of autonomous driving models and enables efficient data pruning with low computational overhead. This facilitates its application to large-scale datasets and real-time processing on vehicle platforms. Experimental results on the NuPlan dataset validate the effectiveness of our method, demonstrating that it can prune up to 40\% of the data without degrading model performance. This offers a practical solution for reducing storage demands while maintaining the functional integrity of autonomous driving systems. Although the scope of our experiments is relatively limited, we hope this work contributes valuable insights toward scalable data pruning strategies in the field of autonomous driving. 





\bibliography{refs}          

@misc{xu2025wode2ewaymoopendataset,
      title={WOD-E2E: Waymo Open Dataset for End-to-End Driving in Challenging Long-tail Scenarios}, 
      author={Runsheng Xu and Hubert Lin and Wonseok Jeon and Hao Feng and Yuliang Zou and Liting Sun and John Gorman and Kate Tolstaya and Sarah Tang and Brandyn White and Ben Sapp and Mingxing Tan and Jyh-Jing Hwang and Dragomir Anguelov},
      year={2025},
      eprint={2510.26125},
      archivePrefix={arXiv},
      primaryClass={cs.CV},
      url={https://arxiv.org/abs/2510.26125}, 
}

@article{li2023survey,
  title={A survey on self-evolving autonomous driving: a perspective on data closed-loop technology},
  author={Li, Xincheng and Wang, Zhaoyi and Huang, Yanjun and Chen, Hong},
  journal={IEEE Transactions on Intelligent Vehicles},
  volume={8},
  number={11},
  pages={4613--4631},
  year={2023},
  publisher={IEEE}
}

@InProceedings{pmlr-v97-ghorbani19c,
  title = 	 {Data Shapley: Equitable Valuation of Data for Machine Learning},
  author =       {Ghorbani, Amirata and Zou, James},
  booktitle = 	 {Proceedings of the 36th International Conference on Machine Learning},
  pages = 	 {2242--2251},
  year = 	 {2019},
  editor = 	 {Chaudhuri, Kamalika and Salakhutdinov, Ruslan},
  volume = 	 {97},
  series = 	 {Proceedings of Machine Learning Research},
  month = 	 {09--15 Jun},
  publisher =    {PMLR},
  url = 	 {https://proceedings.mlr.press/v97/ghorbani19c.html}
}

@article{mahmud2017application,
  title={Application of proximal surrogate indicators for safety evaluation: A review of recent developments and research needs},
  author={Mahmud, SM Sohel and Ferreira, Luis and Hoque, Md Shamsul and Tavassoli, Ahmad},
  journal={IATSS research},
  volume={41},
  number={4},
  pages={153--163},
  year={2017},
  publisher={Elsevier}
}

@article{tian2024drivevlm,
  title={Drivevlm: The convergence of autonomous driving and large vision-language models},
  author={Tian, Xiaoyu and Gu, Junru and Li, Bailin and Liu, Yicheng and Wang, Yang and Zhao, Zhiyong and Zhan, Kun and Jia, Peng and Lang, Xianpeng and Zhao, Hang},
  journal={arXiv preprint arXiv:2402.12289},
  year={2024}
}

@article{wang2024survey,
  title={A survey on datasets for the decision making of autonomous vehicles},
  author={Wang, Yuning and Han, Zeyu and Xing, Yining and Xu, Shaobing and Wang, Jianqiang},
  journal={IEEE Intelligent Transportation Systems Magazine},
  volume={16},
  number={2},
  pages={23--40},
  year={2024},
  publisher={IEEE}
}

@article{cheng2024pluto,
  title={Pluto: Pushing the limit of imitation learning-based planning for autonomous driving},
  author={Cheng, Jie and Chen, Yingbing and Chen, Qifeng},
  journal={arXiv preprint arXiv:2404.14327},
  year={2024}
}

@article{zhou2022dynamically,
  title={Dynamically conservative self-driving planner for long-tail cases},
  author={Zhou, Weitao and Cao, Zhong and Deng, Nanshan and Liu, Xiaoyu and Jiang, Kun and Yang, Diange},
  journal={IEEE Transactions on Intelligent Transportation Systems},
  volume={24},
  number={3},
  pages={3476--3488},
  year={2022},
  publisher={IEEE}
}

@article{zhang2023perception,
  title={Perception and sensing for autonomous vehicles under adverse weather conditions: A survey},
  author={Zhang, Yuxiao and Carballo, Alexander and Yang, Hanting and Takeda, Kazuya},
  journal={ISPRS Journal of Photogrammetry and Remote Sensing},
  volume={196},
  pages={146--177},
  year={2023},
  publisher={Elsevier}
}

@article{paul2021deep,
  title={Deep learning on a data diet: Finding important examples early in training},
  author={Paul, Mansheej and Ganguli, Surya and Dziugaite, Gintare Karolina},
  journal={Advances in neural information processing systems},
  volume={34},
  pages={20596--20607},
  year={2021}
}

@article{yang2022dataset,
  title={Dataset pruning: Reducing training data by examining generalization influence},
  author={Yang, Shuo and Xie, Zeke and Peng, Hanyu and Xu, Min and Sun, Mingming and Li, Ping},
  journal={arXiv preprint arXiv:2205.09329},
  year={2022}
}

@article{toneva2018empirical,
  title={An empirical study of example forgetting during deep neural network learning},
  author={Toneva, Mariya and Sordoni, Alessandro and Combes, Remi Tachet des and Trischler, Adam and Bengio, Yoshua and Gordon, Geoffrey J},
  journal={arXiv preprint arXiv:1812.05159},
  year={2018}
}

@article{sorscher2022beyond,
  title={Beyond neural scaling laws: beating power law scaling via data pruning},
  author={Sorscher, Ben and Geirhos, Robert and Shekhar, Shashank and Ganguli, Surya and Morcos, Ari},
  journal={Advances in Neural Information Processing Systems},
  volume={35},
  pages={19523--19536},
  year={2022}
}

@article{hao2025driveaction,
  title={Driveaction: A benchmark for exploring human-like driving decisions in vla models},
  author={Hao, Yuhan and Li, Zhengning and Sun, Lei and Wang, Weilong and Yi, Naixin and Song, Sheng and Qin, Caihong and Zhou, Mofan and Zhan, Yifei and Lang, Xianpeng},
  journal={arXiv preprint arXiv:2506.05667},
  year={2025}
}

@inproceedings{hallgarten2024can,
  title={Can vehicle motion planning generalize to realistic long-tail scenarios?},
  author={Hallgarten, Marcel and Zapata, Julian and Stoll, Martin and Renz, Katrin and Zell, Andreas},
  booktitle={2024 IEEE/RSJ International Conference on Intelligent Robots and Systems (IROS)},
  pages={5388--5395},
  year={2024},
  organization={IEEE}
}

@article{caesar2021nuplan,
  title={nuplan: A closed-loop ml-based planning benchmark for autonomous vehicles},
  author={Caesar, Holger and Kabzan, Juraj and Tan, Kok Seang and Fong, Whye Kit and Wolff, Eric and Lang, Alex and Fletcher, Luke and Beijbom, Oscar and Omari, Sammy},
  journal={arXiv preprint arXiv:2106.11810},
  year={2021}
}

@inproceedings{evans2024bad,
  title={Bad students make great teachers: Active learning accelerates large-scale visual understanding},
  author={Evans, Talfan and Pathak, Shreya and Merzic, Hamza and Schwarz, Jonathan and Tanno, Ryutaro and Henaff, Olivier J},
  booktitle={European Conference on Computer Vision},
  pages={264--280},
  year={2024},
  organization={Springer}
}

@inproceedings{qian2024spider,
  title={SPIDER: Self-Driving Planners and Intelligent Decision-Making Engines with Reusability},
  author={Qian, Zelin and Jiang, Kun and Cao, Zhong and Qian, Kangan and Xu, Yiliang and Zhou, Weitao and Yang, Diange},
  booktitle={2024 IEEE 27th International Conference on Intelligent Transportation Systems (ITSC)},
  pages={937--944},
  year={2024},
  organization={IEEE}
}

@article{zhou2025drarl,
  title={DRARL: Disengagement-Reason-Augmented Reinforcement Learning for Efficient Improvement of Autonomous Driving Policy},
  author={Zhou, Weitao and Zhang, Bo and Cao, Zhong and Li, Xiang and Cheng, Qian and Liu, Chunyang and Zhang, Yaqin and Yang, Diange},
  journal={arXiv preprint arXiv:2506.16720},
  year={2025}
}

@article{wang2018dataset,
  title={Dataset distillation},
  author={Wang, Tongzhou and Zhu, Jun-Yan and Torralba, Antonio and Efros, Alexei A},
  journal={arXiv preprint arXiv:1811.10959},
  year={2018}
}

@inproceedings{cazenavette2022dataset,
  title={Dataset distillation by matching training trajectories},
  author={Cazenavette, George and Wang, Tongzhou and Torralba, Antonio and Efros, Alexei A and Zhu, Jun-Yan},
  booktitle={Proceedings of the IEEE/CVF Conference on Computer Vision and Pattern Recognition},
  pages={4750--4759},
  year={2022}
}

@article{cao2023continuous,
  title={Continuous improvement of self-driving cars using dynamic confidence-aware reinforcement learning},
  author={Cao, Zhong and Jiang, Kun and Zhou, Weitao and Xu, Shaobing and Peng, Huei and Yang, Diange},
  journal={Nature Machine Intelligence},
  volume={5},
  number={2},
  pages={145--158},
  year={2023},
  publisher={Nature Publishing Group UK London}
}

@inproceedings{zhou2022long,
  title={Long-tail prediction uncertainty aware trajectory planning for self-driving vehicles},
  author={Zhou, Weitao and Cao, Zhong and Xu, Yunkang and Deng, Nanshan and Liu, Xiaoyu and Jiang, Kun and Yang, Diange},
  booktitle={2022 IEEE 25th International Conference on Intelligent Transportation Systems (ITSC)},
  pages={1275--1282},
  year={2022},
  organization={IEEE}
}

@article{yin2023squeeze,
  title={Squeeze, recover and relabel: Dataset condensation at imagenet scale from a new perspective},
  author={Yin, Zeyuan and Xing, Eric and Shen, Zhiqiang},
  journal={Advances in Neural Information Processing Systems},
  volume={36},
  pages={73582--73603},
  year={2023}
}

@article{kaplan2020scaling,
  title={Scaling laws for neural language models},
  author={Kaplan, Jared and McCandlish, Sam and Henighan, Tom and Brown, Tom B and Chess, Benjamin and Child, Rewon and Gray, Scott and Radford, Alec and Wu, Jeffrey and Amodei, Dario},
  journal={arXiv preprint arXiv:2001.08361},
  year={2020}
}

@article{berrueta2024maximum,
  title={Maximum diffusion reinforcement learning},
  author={Berrueta, Thomas A and Pinosky, Allison and Murphey, Todd D},
  journal={Nature Machine Intelligence},
  volume={6},
  number={5},
  pages={504--514},
  year={2024},
  publisher={Nature Publishing Group UK London}
}


\end{document}